\documentclass[conference]{IEEEtran}
\IEEEoverridecommandlockouts
\usepackage{cite}
\usepackage{amsmath,amssymb,amsfonts}
\usepackage{algorithmic}
\usepackage{graphicx}
\usepackage{textcomp}
\usepackage{xcolor}
\usepackage{tabularx}
\usepackage{siunitx}
\usepackage{comment}
\usepackage{hyperref}

\usepackage{multirow}
\usepackage{multicol}

\usepackage{cleveref}
\crefname{table}{TABLE}{TABLEs}
\Crefname{table}{TABLE}{TABLEs}
\crefname{figure}{Fig.}{Figs.}
\Crefname{figure}{Fig.}{Figs.}
\crefname{section}{Section}{Section}
\Crefname{section}{Section}{Section}
\usepackage{svg}
\usepackage[export]{adjustbox}
\usepackage{subcaption}
\usepackage{caption}
\newcommand{\RotTxt}[1]{\rotatebox[origin=c]{90}{\parbox[c]{1cm}{\centering #1}}}

\def\BibTeX{{\rm B\kern-.05em{\sc i\kern-.025em b}\kern-.08em
    T\kern-.1667em\lower.7ex\hbox{E}\kern-.125emX}}
\begin{document}

    \title{Medical Image Retrieval Using Pretrained Embeddings 
}

\author{\IEEEauthorblockN{ Farnaz Khun Jush}
\IEEEauthorblockA{\textit{Bayer AG}, Berlin, Germany \\
farnaz.khunjush@bayer.com} 
\and
\IEEEauthorblockN{Tuan Truong}
\IEEEauthorblockA{\textit{Bayer AG}, Berlin, Germany \\
tuan.truong@bayer.com}
\and
\IEEEauthorblockN{Steffen Vogler}
\IEEEauthorblockA{\textit{Bayer AG}, Berlin, Germany \\
steffen.vogler@bayer.com}
\and
\IEEEauthorblockN{Matthias Lenga}
\IEEEauthorblockA{\textit{Bayer AG}, Berlin, Germany\\
matthias.lenga@bayer.com}
}


\maketitle

\begin{abstract}

A wide range of imaging techniques and data formats available for medical images make accurate retrieval from image databases challenging. 
 Efficient retrieval systems are crucial in advancing medical research, enabling large-scale studies and innovative diagnostic tools. 
Thus, addressing the challenges of medical image retrieval is essential for the continued enhancement of healthcare and research. 
 In this study, we evaluated the feasibility of employing four state-of-the-art pretrained models for medical image retrieval at modality, body region, and organ levels and compared the results of two similarity indexing approaches. Since the employed networks take 2D images, we analyzed the impacts of weighting and sampling strategies to incorporate 3D information during retrieval of 3D volumes. 
We showed that medical image retrieval is feasible using pretrained networks without any additional training or fine-tuning steps. Using pretrained embeddings, we achieved a recall of 1 for various tasks at modality, body region, and organ level. 
 
\end{abstract}


\section{Introduction}

In the field of computer vision, researchers have been studying content-based image retrieval (CBIR) for the past decades \cite{dubey2021decade}. CBIR systems typically store low-dimensional representations of images in a database and retrieve similar images based on representation similarity. Traditionally, manual crafting of distinctive features led to a semantic gap, where important image details were lost due to low-dimensional feature design \cite{wang2022two, dubey2021decade}. Recent advances in deep learning have shifted the focus toward machine-generated discriminative feature spaces, to effectively bridge this semantic gap.

While natural image retrieval has seen extensive research, applying retrieval frameworks to medical images, especially radiology images, remains challenging. Previous efforts often relied on manually crafted or shallow learning-based features for various medical image modalities, which are insufficient for large-scale databases \cite{haq2021deep}. Deep learning-based approaches show promise, but the lack of large-scale radiology datasets has been a hurdle \cite{wickstrom2023clinically}. 
There are several pretrained models that are trained on huge publicly available image datasets for various tasks, e.g., image segmentation and classification. These datasets often consist of natural images e.g., ImageNet \cite{russakovsky2015imagenet} or COCO \cite{lin2014microsoft}. 
It was shown that in a supervised training approach domain-specific pretraining gives rise to improved performance in downstream tasks compared to the models trained on natural images \cite{mei2022radimagenet}. 
Additionally, pretrained models using annotated medical images demonstrated better interpretability compared to the models trained using ImageNet, especially for smaller radiology datasets \cite{mei2022radimagenet}.

On the other hand, studies showed that ImageNet pretrained features are in principle transferable to medical imaging tasks \cite{truong2021transferable} e.g., feature extractors using knowledge distillation with vision transformer (ViT) \cite{dosovitskiy2020image} backbones like DINO \cite{caron2021emerging} yield richer embeddings on the downstream tasks than supervised pretrained models \cite{truong2021transferable}. 

To the best of our knowledge, no study has assessed the feasibility and performance of utilizing pretrained embeddings, originally trained for non-CBIR tasks, in the context of medical image retrieval.
The contribution of this study is as follows: 

\begin{itemize}
    \item We evaluated the performance of using start-of-the-art \textbf{self-supervised pretrained models} i.e., DINOv1 \cite{caron2021emerging}, DINOv2 \cite{oquab2023dinov2}, DreamSim \cite{fu2023dreamsim} trained on natural images versus a \textbf{supervised pretrained model} based on Swin Transformer  \cite{liu2021swin} and a ResNet50 model \cite{he2016deep} trained on RadImageNet dataset presented in \cite{mei2022radimagenet} for 3D medical image retrieval tasks at the modality, body region, and organ level. 
    \item We compared the results of two different vector database indexing approaches (LSH and HNSW) for medical image retrieval in combination with different embedding types.
    \item Since all the used networks are trained on 2D images and medical images are often 3D volumes, aggregating the information from various slides plays a crucial role in the retrieval of 3D images. We propose a simple, yet effective count-based scheme that allows aggregating 2D similarity information across the 3D image volumes. Moreover, we compare the impact of weighting and different sub-sampling techniques on the overall performance.
\end{itemize}

\begin{figure*}[!htpb]
  \centering
  \includegraphics[trim=2 10.0cm .5cm 0, clip, scale=0.58]{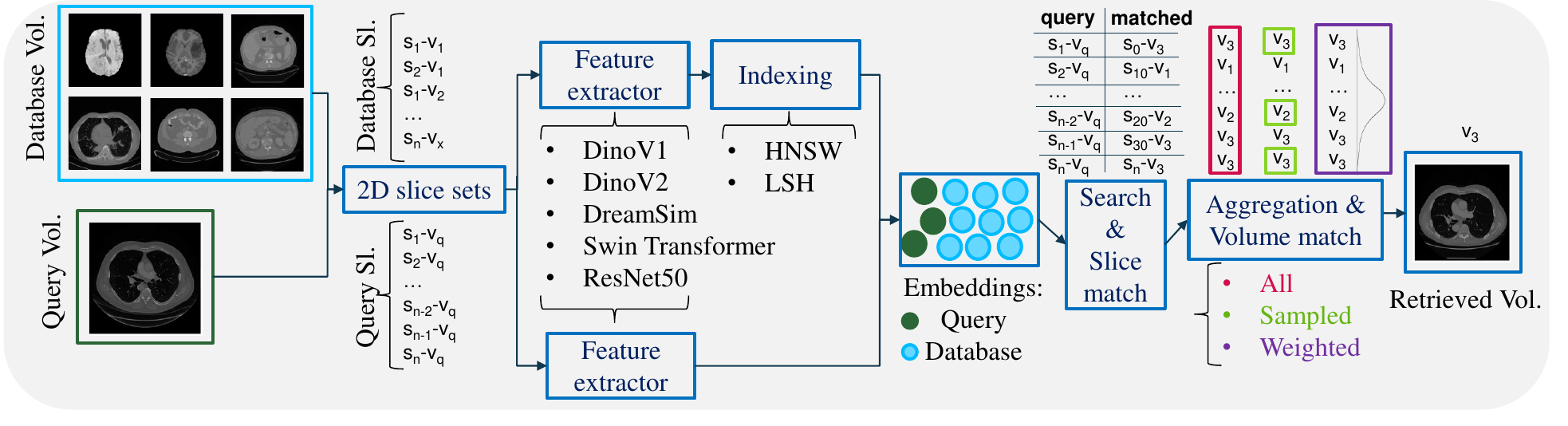}
  \caption{Overview of the proposed content-based 3D medical image retrieval system and ablation experiments. Note: $s_x-v_y$ refers to slice (Sl.) $x$ of volume (Vol.) $y$.}
  \label{fig: overview}
\end{figure*}

\cref{fig: overview} presents a schematic overview of the proposed method for medical image retrieval.

\section{Methods}
\subsection{Vector Database and Indexing}

In the context of image search, the images and related metadata, e.g. annotations, are typically stored and organized in a database.
In contrast, a query represents the user or system's specific request for images, which can generally be in the form of a reference image itself, a textual description, or a combination of both.
The primary goal of a query is to search the image database for relevant matches based on criteria such as content or metadata. 
In the scope of this work, the search process involves comparing a query image to all images within the database to find the most similar image based on the minimum distance of the embeddings. We do not rely on any metadata information in any step. 

A similarity search is used to compare data quickly and return relevant results. As the database size grows, the challenges in accuracy and speed increase, making naive approaches ineffective. To address this, vector representations known as embeddings are created using statistical, machine learning, or deep learning techniques. These embeddings are indexed for efficient vector similarity search. 
This study compared two indexing approaches: Locality Sensitive Hashing (LSH) \cite{charikar2002similarity} and Hierarchical Navigable Small World (HNSW) \cite{malkov2018efficient}, in the context of medical image retrieval. LSH hashes data points to improve search efficiency, while HNSW organizes data into a hierarchical graph structure for efficient navigation during search, especially suitable for larger datasets with high dimensionality. We used the Facebook AI Similarity Search (FAISS) package that offers state-of-the art implementations of the datastructures mentioned above \cite{johnson2019billion}.

\subsection{Feature Extraction}
The performance of four different slice embedding extractors is benchmarked for the CBIR tasks. We use DINOv1 \cite{caron2021emerging}, DINOv2 \cite{oquab2023dinov2} and DreamSim \cite{fu2023dreamsim} which were pretrained on ImageNet in a \emph{self-supervised} fashion. DINO was the first model of this series published \cite{caron2021emerging}, and could learn efficient representations from unlabeled data using self-distillation \cite{caron2021emerging, he2020momentum}. 
DreamSim is based on DINOv1 and is fine-tuned on synthetic images and a new perceptual metric that aligns with human perception \cite{fu2023dreamsim}.
DINOv2 revisited existing approaches and combined various techniques to scale the pretraining process in terms of both data and model size \cite{oquab2023dinov2} which outperformed DINOv1 in various tasks at the time of its publication. 
%
Furthermore, a Swin Transformer model \cite{liu2021swin} and a ResNet50 model \cite{he2016deep} which are trained \emph{supervised} on the RadImageNet \cite{mei2022radimagenet} is used.

\subsection{Dataset}

We designed a CBIR benchmark relying on the Medical Segmentation Decathlon (MSD) challenge dataset which is publicly available \cite{antonelli2022medical}. The composition of the original dataset is shown in the \cref{tab: MSD Dataset}. 
The query cases are sourced from the MSD Challenge test split, while the train set serves as the database for searching.
The search is assessed on three different levels: \emph{modality}, \emph{body regions}, and \emph{organs}. The modality category consists of CT or MR scans (protocols are not included). 
Organ labels and body regions mapping are shown in \cref{tab: MSD Dataset}. 
The organ level labels used for the evaluation were derived from the associated MSD tasks. Thus, the organ level label of a case refers to just one of the existing organs in the image.

\begin{table}[!htpb]
    \centering
    \tiny { 
    \caption{MSD Challenge Dataset}
    \label{tab: MSD Dataset}
    \begin{tabular}{|c|c|c|c|l|c|}
        \hline
        Organ & Modality  & Cases (Train / Test) & Body region\\
        \hline
        Brain (BRATS) & MRI / 4D & 732 (466 / 266) & Head\\
        \hline
        Colon & CT / 3D & 190 (126 / 64) & Abdomen\\
        \hline
        Hepatic vessels & CT / 3D &  443 (303 / 140) & Abdomen \\
        \hline
        Hippocampus & MRI / 3D &  390 (260 / 130)& Head \\
        \hline
        Heart (La) & MRI / 3D &30 (20 / 10) & Chest \\
        \hline
        Liver & CT / 3D &201 (131 / 70) & Abdomen\\
        \hline
        Lung & CT / 3D &95 (63 / 32) & Chest\\
        \hline
        Pancreas & CT / 3D &  420 (281 / 139) & Abdomen \\
        \hline
        Prostate & MRI / 4D & 48 (32 / 16) & Pelvis \\
        \hline
        Spleen & CT / 3D &  61 (41 / 20) & Abdomen\\
        \hline
    \end{tabular}
    }
\end{table}

MSD challenge dataset as shown in \cref{tab: MSD Dataset} consists of 3D and 4D volumes (3D volumes with multiple protocols). 
In the case of multiple protocols, first 3D volumes are extracted, and then individual 2D slices of the corresponding 3D volumes are utilized for embedding extraction.
The input size for all the used models is equal to $224\times224$ pixels with image replication along the RGB channel axis. 
For all the ViT-based models images are normalized to the ImageNet mean and standard deviation of $(0.485, 0.456, 0.406)$ and $(0.229, 0.224, 0.225)$, respectively. 
For the ResNet50 model, the images are normalized to $0.5$ mean and $0.5$ standard deviation based on \cite{mei2022radimagenet}. 
The total size of our database is $439,782$ embeddings, while the final query set comprises $237,883$ embeddings.

\begin{table*}[!t]
    \centering
     {\tiny
    \caption{Recall for different body regions Abdomen, Chest, Head, and Pelvis. Comparison: different vector database indexes (Id.), sampling strategies, aggregation variants all, per Image-id level and slice wise (SW)}
    \label{tab:Body region}
    \begin{tabular}{|p{0.2cm}|p{0.2cm}|p{0.8cm}|p{0.4cm}p{0.4cm}p{0.4cm}p{0.4cm}p{0.6cm}|p{0.4cm}p{0.4cm}p{0.4cm}p{0.4cm}p{0.4cm}p{0.6cm}|p{0.4cm}p{0.4cm}p{0.6cm}|p{0.6cm}|p{0.6cm}|p{0.6cm}|p{0.6cm}|} \hline
    \multirow{2}{*}{} & \multirow{2}{*}{Id.}  & \multirow{2}{*}{Embedding}  & \multicolumn{5}{|c|}{\# Random Samples}  & \multicolumn{6}{|c|}{\# Sampled every x mm} & \multicolumn{3}{|c|}{\# every n slices} & \multirow{2}{*}{all} & \multirow{2}{*}{W} & \multirow{2}{*}{SW}\\ \cline{4-17} 
         & & & 5 & 10 & 20 & 40 & 80 & 3 & 5 & 7 & 10 & 12 & 15 & 3 & 5 & 10 & & &\\ \hline \cline{1-17}
    \multirow{10}{*}{\RotTxt{Abdomen}}     & \multirow{5}{*}{\RotTxt{HNSW}} &  DreamSim & .993 & .990&.995 &.988 &.997 & .993& .993& .993& .993& .993& .995&.993 & .995 &.995 & .993& \textbf{1.0} &.983\\ 
         &  &  DINOv1 &  .857&  .880& .880& .880& .995&  \textbf{1.0} & \textbf{1.0} & \textbf{1.0} & .997 & .997  &\textbf{1.0}& \textbf{1.0}& \textbf{1.0}&.997 &\textbf{1.0}&\textbf{1.0}&.988\\
         &  &  DINOv2 & .988 & .990 & .997& .993& .995& .993& .993& .993& .993&.993 &.990 &.993 & .995& .995 &.993& \textbf{1.0} &.979\\ 
         &  &  ResNet50 & .981& .988& .990& .988& .990& .990&.990 &.990 & .990& .990& .990& .990&.990 & .993& .990& \textbf{1.0}&  .976 \\ 
         &  & SwinTrans.& .983 & .995 & .993 & .995 & .993 & .995 & .993 & .995 & .995 & .993 &  .997 & .995 & \textbf{1.0}  &  \textbf{1.0} & .997& .995 &  .987 \\ \cline{2-3}
         & \multirow{5}{*}{\RotTxt{LSH}} &  DreamSim & .993 & .995 & .990&.993 & .995& .995& .995&.993 & \textbf{.997}& \textbf{.997} &.990 &.993 &.995&.995 &.993&\textbf{.997}&.981\\
         &  &  DINOv1 & .995 &  \textbf{1.0}&.997 &.995 &  \textbf{1.0}&\textbf{1.0} & \textbf{1.0}&\textbf{1.0} & \textbf{1.0}& \textbf{1.0}&\textbf{1.0} & \textbf{1.0}& \textbf{1.0}&\textbf{1.0} &\textbf{1.0}&\textbf{1.0}&.986\\
         &  &  DINOv2 & .988 & .995 &.993 &.988 &.993 & .993& .993& .993& .993& .995&.995 &.995 &.995 &.995 &.993&\textbf{1.0}&.974\\ 
         &  &  ResNet50 & .983& .981& .983& .988& .988& .990 &.990 & .993& \textbf{.997}& .995& .995&  .993& \textbf{.997}&.986 &.990 & \textbf{.997} & .979  \\ 
         &  & SwinTrans.& \textbf{.997 }& .986 & .995 & .995 & \textbf{.997} &  \textbf{.997} & .995 & .995 & .995 & .995 & .993 &  .995 & \textbf{.997} & \textbf{.997}  &\textbf{.997} & .995 & .985 \\ \hline
    \multirow{10}{*}{\RotTxt{Chest}}   & \multirow{5}{*}{\RotTxt{HNSW}}  &  DreamSim & .857 &  .857&.809 & .857&.857 & \textbf{.904}&.857 & .833&.857 & .857& .785&.880 &.880 &.833 &\textbf{.904}&.595&.738\\ 
         &  &  DINOv1 &  .857& .880 & .880& .880&.880 &.904 &.904 & \textbf{.928}&\textbf{.928} &.880 &.857 & .904 & .904 & .880 & .904&.690&.804\\
         &  &  DINOv2 & .809 & .833 &\textbf{.880 }&.857 &\textbf{.880 }&.857 &\textbf{.880 }&.857 & \textbf{.880} &.833 &.833 &.857 & \textbf{.880 }&\textbf{.880 }& \textbf{.880} &.619&.742\\ 
         &  &  ResNet50 & .880& .880& .880& .928& .928&  .904& .928& .904& .928& .904& .904 & .904 & \textbf{.952} & .928&.904 &.547 &  .744 \\
         &  & SwinTrans.  & .880 & .880 & \textbf{.952} & .880 & .928 & .928 & \textbf{.952 }& .904 & .992 & .928 & \textbf{.952} &  .928 & .904 & .904 &.928 & .761&  .822\\ \cline{2-3}
         & \multirow{4}{*}{\RotTxt{LSH}} &  DreamSim & .809 & .880 &.833 & .809&.833 &  .857 & \textbf{.904}  & .883 & .880 & .880 &  .880& .880 & .857&.857 &.857&.547&.725\\ 
         &  &  DINOv1 &  .809&  .857&.880 &\textbf{.904} &.880 & .880&.880 &\textbf{.904} &.857 &.880  &.857 &.880 &\textbf{.904} &.857&\textbf{.904}&.690&.768\\
         &  &  DINOv2 & .738 & .833 &.833 & .833& .833&.857 &.857 & \textbf{.880} &.833 &.833 & .833&.833 & \textbf{.880} & .833 & .857 & .5 & .720\\
         &  &  ResNet50 & .857& .857& .857& .857& .904&.904 & .976& \textbf{.976}& .928& .904&.928 & .928& \textbf{.976}& .928& \textbf{.976}&.547 &  .741 \\
         &  & SwinTrans.& .880 & .833 & \textbf{.928} & .904 & .857 & .880 & .857 & .880 & .833 & .904 & .904 & .880 & .904 & .857 & .880 & .761 &  .803 \\ \hline
    \multirow{10}{*}{\RotTxt{Head}}     & \multirow{5}{*}{\RotTxt{HNSW}} &  DreamSim & \textbf{1.0} & \textbf{1.0} & \textbf{1.0}&\textbf{1.0} & \textbf{1.0}&\textbf{1.0} & \textbf{1.0}&\textbf{1.0} &\textbf{1.0} &\textbf{1.0} & \textbf{1.0}& \textbf{1.0}& \textbf{1.0}&\textbf{1.0}&\textbf{1.0}&\textbf{1.0}&\textbf{1.0}\\ 
         &  &  DINOv1 & \textbf{1.0} & \textbf{1.0} & \textbf{1.0}&\textbf{1.0} & \textbf{1.0}&\textbf{1.0} & \textbf{1.0}&\textbf{1.0} &\textbf{1.0} &\textbf{1.0} & \textbf{1.0}& \textbf{1.0}& \textbf{1.0}&\textbf{1.0}&\textbf{1.0}&\textbf{1.0}&\textbf{1.0}\\ 
         &  &  DINOv2 & \textbf{1.0} &\textbf{1.0}  & \textbf{1.0}& \textbf{1.0}&\textbf{1.0} & \textbf{1.0}& \textbf{1.0}& \textbf{1.0}& .997& \textbf{1.0}& \textbf{1.0}& \textbf{1.0}& \textbf{1.0}&.997&\textbf{1.0}&\textbf{1.0}&.99\\  
         &  &  ResNet50 & \textbf{1.0}& \textbf{1.0}& \textbf{1.0}& \textbf{1.0}&\textbf{1.0} & \textbf{1.0}&\textbf{1.0} &\textbf{1.0} &\textbf{1.0} &\textbf{1.0} &\textbf{1.0} &\textbf{1.0} &\textbf{1.0} &\textbf{1.0} &\textbf{1.0} &\textbf{1.0} &  \textbf{1.0} \\ 
           &  & SwinTrans.& \textbf{1.0}& \textbf{1.0}& \textbf{1.0}& \textbf{1.0}&\textbf{1.0} & \textbf{1.0}&\textbf{1.0} &\textbf{1.0} &\textbf{1.0} &\textbf{1.0} &\textbf{1.0} &\textbf{1.0} &\textbf{1.0} &\textbf{1.0} &\textbf{1.0} &\textbf{1.0} &  \textbf{1.0} \\ \cline{2-3}
         & \multirow{5}{*}{\RotTxt{LSH}} &  DreamSim & \textbf{1.0} & \textbf{1.0} & \textbf{1.0}& \textbf{1.0}& \textbf{1.0}&\textbf{1.0} & \textbf{1.0}& \textbf{1.0}&\textbf{1.0} &\textbf{1.0} &\textbf{1.0} & \textbf{1.0}& \textbf{1.0}&\textbf{1.0}&\textbf{1.0}&\textbf{1.0}&\textbf{1.0}\\ 
         &  &  DINOv1 & \textbf{1.0} &\textbf{1.0}  &\textbf{1.0} &\textbf{1.0} &\textbf{1.0} & \textbf{1.0}& \textbf{1.0}&\textbf{1.0} & \textbf{1.0}&\textbf{1.0} & \textbf{1.0}&\textbf{1.0} & \textbf{1.0}&\textbf{1.0}&\textbf{1.0}&\textbf{1.0}&.99\\
         &  &  DINOv2 & \textbf{1.0} &\textbf{1.0}  & \textbf{1.0}&\textbf{1.0} & \textbf{1.0}& \textbf{1.0}& \textbf{1.0}& \textbf{1.0}&\textbf{1.0} &\textbf{1.0} &\textbf{1.0} &\textbf{1.0} &\textbf{1.0} &\textbf{1.0}&\textbf{1.0}&\textbf{1.0}&.99\\ 
         &  &  ResNet50 &\textbf{1.0} & \textbf{1.0}&\textbf{1.0} &\textbf{1.0} & \textbf{1.0}& \textbf{1.0}&\textbf{1.0} &\textbf{1.0} &\textbf{1.0} &\textbf{1.0} &\textbf{1.0} &\textbf{1.0} &\textbf{1.0} & \textbf{1.0}&\textbf{1.0} & \textbf{1.0} &  .999 \\
         &  & SwinTrans.&\textbf{1.0} & \textbf{1.0}&\textbf{1.0} &\textbf{1.0} & \textbf{1.0}& \textbf{1.0}&\textbf{1.0} &\textbf{1.0} &\textbf{1.0} &\textbf{1.0} &\textbf{1.0} &\textbf{1.0} &\textbf{1.0} & \textbf{1.0}&\textbf{1.0} & \textbf{1.0} &  .999 \\ \hline
    \multirow{10}{*}{\RotTxt{Pelvis}} & \multirow{5}{*}{\RotTxt{HNSW}} &  DreamSim &\textbf{1.0}  & \textbf{1.0} &\textbf{1.0} &\textbf{1.0} &\textbf{1.0} &\textbf{1.0} & \textbf{1.0}&\textbf{1.0} &\textbf{1.0} &\textbf{1.0} & \textbf{1.0}& \textbf{1.0}& \textbf{1.0}&\textbf{1.0}&\textbf{1.0}&.937&.977\\ 
         &  &  DINOv1 & .866 & .75 &.812 &.769 &.687 & .812& .812&.812 &.714 &.875 &.928 &.875 & \textbf{1.0}&\textbf{1.0}&.733&\textbf{1.0}&.846\\ 
         &  &  DINOv2 & \textbf{1.0} & \textbf{1.0} &\textbf{1.0} &\textbf{1.0} & \textbf{1.0}& \textbf{1.0}& \textbf{1.0}& \textbf{1.0}& \textbf{1.0}& \textbf{1.0}&\textbf{1.0}  &\textbf{1.0} &\textbf{1.0} &\textbf{1.0}&\textbf{1.0}&\textbf{1.0}&.980\\  
         &  &  ResNet50 & \textbf{1.0}& \textbf{1.0}& \textbf{1.0}& \textbf{1.0}& \textbf{1.0}& \textbf{1.0}&\textbf{1.0} &\textbf{1.0} &\textbf{1.0} &\textbf{1.0} &\textbf{1.0} & \textbf{1.0}& \textbf{1.0}&\textbf{1.0} &\textbf{1.0} & \textbf{1.0}& .979 \\
         &  & SwinTrans. & \textbf{1.0}& \textbf{1.0}& \textbf{1.0}& \textbf{1.0}& \textbf{1.0}& \textbf{1.0}&\textbf{1.0} &\textbf{1.0} &\textbf{1.0} &\textbf{1.0} &\textbf{1.0} & \textbf{1.0}& \textbf{1.0}&\textbf{1.0} &\textbf{1.0} & \textbf{1.0}& .984 \\ \cline{2-3}
         & \multirow{5}{*}{\RotTxt{LSH}}  &  DreamSim &  \textbf{1.0}& \textbf{1.0} & \textbf{1.0}&\textbf{1.0} &\textbf{1.0} & \textbf{1.0}&\textbf{1.0} &\textbf{1.0} &\textbf{1.0} & \textbf{1.0}&\textbf{1.0} &\textbf{1.0} &\textbf{1.0} &\textbf{1.0}&\textbf{1.0}&\textbf{1.0}&.986\\ 
         &  &  DINOv1 & \textbf{1.0} & \textbf{1.0} &\textbf{1.0} &\textbf{1.0} &\textbf{1.0} & \textbf{1.0}& \textbf{1.0}&\textbf{1.0} &\textbf{1.0} &\textbf{1.0} &\textbf{1.0} &\textbf{1.0} & \textbf{1.0}&\textbf{1.0}&\textbf{1.0}&\textbf{1.0}&.986\\ 
 & & DINOv2 &\textbf{1.0} & \textbf{1.0} &\textbf{1.0} &\textbf{1.0} &\textbf{1.0} &\textbf{1.0} & \textbf{1.0}& \textbf{1.0}& \textbf{1.0}& \textbf{1.0}& \textbf{1.0}& \textbf{1.0}&\textbf{1.0} &\textbf{1.0}&\textbf{1.0}&\textbf{1.0}&.986\\
 &  &  ResNet50 &\textbf{1.0} &\textbf{1.0} & \textbf{1.0}&\textbf{1.0} & \textbf{1.0} &\textbf{1.0}&\textbf{1.0} &\textbf{1.0} &\textbf{1.0} &\textbf{1.0} &\textbf{1.0}& \textbf{1.0}& \textbf{1.0}& \textbf{1.0}& \textbf{1.0}&\textbf{1.0} & .986  \\ 
  &  &  SwinTrans. &\textbf{1.0} &\textbf{1.0} & \textbf{1.0}&\textbf{1.0} & \textbf{1.0} &\textbf{1.0}&\textbf{1.0} &\textbf{1.0} &\textbf{1.0} &\textbf{1.0} &\textbf{1.0}& \textbf{1.0}& \textbf{1.0}& \textbf{1.0}& \textbf{1.0}&\textbf{1.0} & .986  \\ \hline
\multicolumn{3}{|c|}{Overall average} & .945 & .951 & .952 & .953 & .957 & .963 & .967 & .966 & .961 & .962 & .961 & .965 & .975 & .967 & .964 & .89 & .986 \\ \hline
    \end{tabular}
    }
\end{table*}

\subsection{Aggregation Strategies}
\label{ref: matching strategies}
The proposed CBIR system relies on pre-trained 2D image embeddings for 3D image retrieval. Given a query case, first 2D slice embeddings are extracted. We propose and benchmark different variants to aggregate 2D information to volume level in order to find the best matching image volume.

\paragraph{All Slices} 
For every 2D slice of a query volume, the most similar slice in the database is retrieved using embedding similarity, and the corresponding volume-ID is stored in a hit table. The volume-ID with the most accumulated hits is then returned as the most similar case in the database, see \cref{fig: overview}.


\paragraph{Weighted Slices}
Since often the border slices of a volume carry less information compared to center slices, a Gaussian weighting scheme can be used to increase the impact of center slices and mitigate the impact of less informative slices. For each slice of the query volume the most similar slice is retrieved from the database, and the corresponding volume-ID is recorded. Then the Gaussian weighting is applied to the matched volume-IDs (meaning, the volume-IDs that are matched for center slices are repeated more than the further matches, thus, their hit score increases). The final result is determined by selecting the volume-ID with the highest weighted hit, see \cref{fig: overview}.

\paragraph{Sampled Slices}
Volumetric medical images contain a high degree of spatial redundancy. Hence,  subsampling of transversal slices can improve the speed of the search without impeding the retrieval performance. For all subsampling strategies, the final result is again obtained based on maximum hit accumulation. The following sampling schemes are compared:

\begin{itemize}
    \item Random subset: For each volume we randomly choose $5, 10 ,20, 40$, or $80$ slices with a uniform distribution.
    \item Equidistant:  We sample a subset of equidistant transversal slices with a gap size of $3, 5 ,7, 10, 12$ or $15$ $mm$.
    \item Fixed steps: the slices are sampled regardless of the spacing of the volume. In this setup, every $3, 5,$ and $10$ slices are sampled. 
\end{itemize}

\section{Results and Discussions}

Here we only report the recall rates, for the precision analysis please refer to Appendix A. The performance analysis is conducted on modality, body region, and organ level, i.e. a case is considered as a True Positive if modality, body region, or organ label of the query and the most similar case match. As a baseline measure, we report the recall of \emph{slice-wise (SW)} matching, i.e. each slice from the query is treated as an individual instance and matched against the most similar slice without aggregation to volume level and performance is evaluated only on slice-wise level.

\subsection{Modality Level}
On the modality level, for all possible combinations of the search index, embedding, and aggregation scheme a recall of $1.0$ was obtained. The slice-wise baseline has a recall of $0.999$ for all the tested variants. Here we observed only a few mismatched black CT and MRI background slices.

\subsection{Body Region Level}
\label{sec: body region}
The recall values on the body region level are shown in \cref{tab:Body region} for all different combinations of search index, embedding, and aggregation scheme.

\subsubsection{All Slices} The Head and Pelvis regions of all the models and indexing approaches have a perfect recall of $1.0$, except for the DINOv1 which shows a dropped performance using HNSW indexing. 

The retrieval of the Chest and Abdomen region is challenging due to the presence of mixed slices of the Abdomen and Chest regions in the volumes. For the retrieval of the Abdomen region DINOv1 has the perfect recall regardless of the indexing. Swin Transformer has a slightly lower recall of $0.997$. DreamSim and DINOv2 perform on par for the Abdomen with a recall of $0.993$. 
For Chest, the best recall belongs to ResNet50 and LSH indexing with a recall of $0.976$. Regardless of the indexing, DINOv1 is the second best with a recall of $0.904$. 
DreamSim performs on par using HNSW but its performance drops when LSH indexing is used. 

\subsubsection{Weighted Slices} A perfect recall of 1.0 for the Head and Pelvis regions for all models (except for DreamSim, HNSW) is achieved. 
The introduction of Gaussian weighting significantly enhances the performance of all models in the Abdomen region, but it comes at the cost of reduced recall in the Chest region. This trade-off is due to the CT volumes often containing overlapping regions. Gaussian weighting accentuates the center slices in Abdomen volumes, reducing the chance of misclassification as Chest cases, even though lung-containing slices are still present. However, for the Chest region, where the lung slices are farther from the center slices, increasing the weight of the center slices raises the likelihood of misclassification as Abdominal volumes.

\subsubsection{Sampled Slices} 
Except for DINOv1 and HNSW indexing, the sampling scenarios perform in most cases on par with using all the slices which is expected due to the spatial redundancy of the medical images. Especially for tasks that are distinct enough, i.e., Head and Pelvis using even a few slices of the volumes can result in perfect retrieval. 
Considering all the body regions, the DINOv1, LSH indexing and sampling every $7$~$mm$ based on the spatial resolution has the best overall performance with the recall of $1.0$ for Abodmen, Head, Pelvis, and the recall of $0.904$ for the Chest region.

\subsection{Organ Level}

In the interest of brevity, we provide the organ level analysis only for the best method identified in the \cref{sec: body region}, i.e. DINOv1 embeddings with LSH search index and $7$mm sub-sampling. The associated confusion matrix on the organ level is shown in \cref{fig: CM figure}. A comprehensive table can be found in Appendix B. 
The recall values of this method are as follows: (Brain, Hippocampus, Heart, Prostate: $1.0$), (Colon, $0.453$), (Hepatic vessel, $0.507$), (Liver, $0.685$), (Lung, $0.875$), (Pancreas, $0.674$), and (Spleen, $0.1$).  \cref{fig: CM figure} shows that the organs belonging to the same body region, e.g. Abdomen are mislabeled as another organ of the Abdomen which is in principle interpretable because the organ level labels are derived from the MSD tasks and often all or more than one of the abdominal organs are visible even on the single slice which justifies for the low recall of the 5 abdomen organs. 
Similarly, CT scans of abdominal and lung organs encompass more than just the specific organ of focus; sometimes, abdominal slices are present in lung CTs, and vice versa, leading to the occasional mislabeling of lung volumes as abdominal organs or abdominal organs as lung volumes. This is considered as label noise that should be taken into account when analyzing the results.

In order to better compare the performance of the different embedding types, we aggregate all recall values obtained from varying the search index and the slice aggregation scheme as discussed in \cref{ref: matching strategies}. \cref{fig:combined max-median} shows per embedding type the median and the maximum recall of the aggregated recall values separately for each organ. The detailed precision results at the organ level were omitted because the values were very close and did not follow any patterns, making their comparison uninformative.

\begin{figure}[!t]
    \centering
    \begin{minipage}{0.48\textwidth}
        \centering
        \includegraphics[scale=0.30]{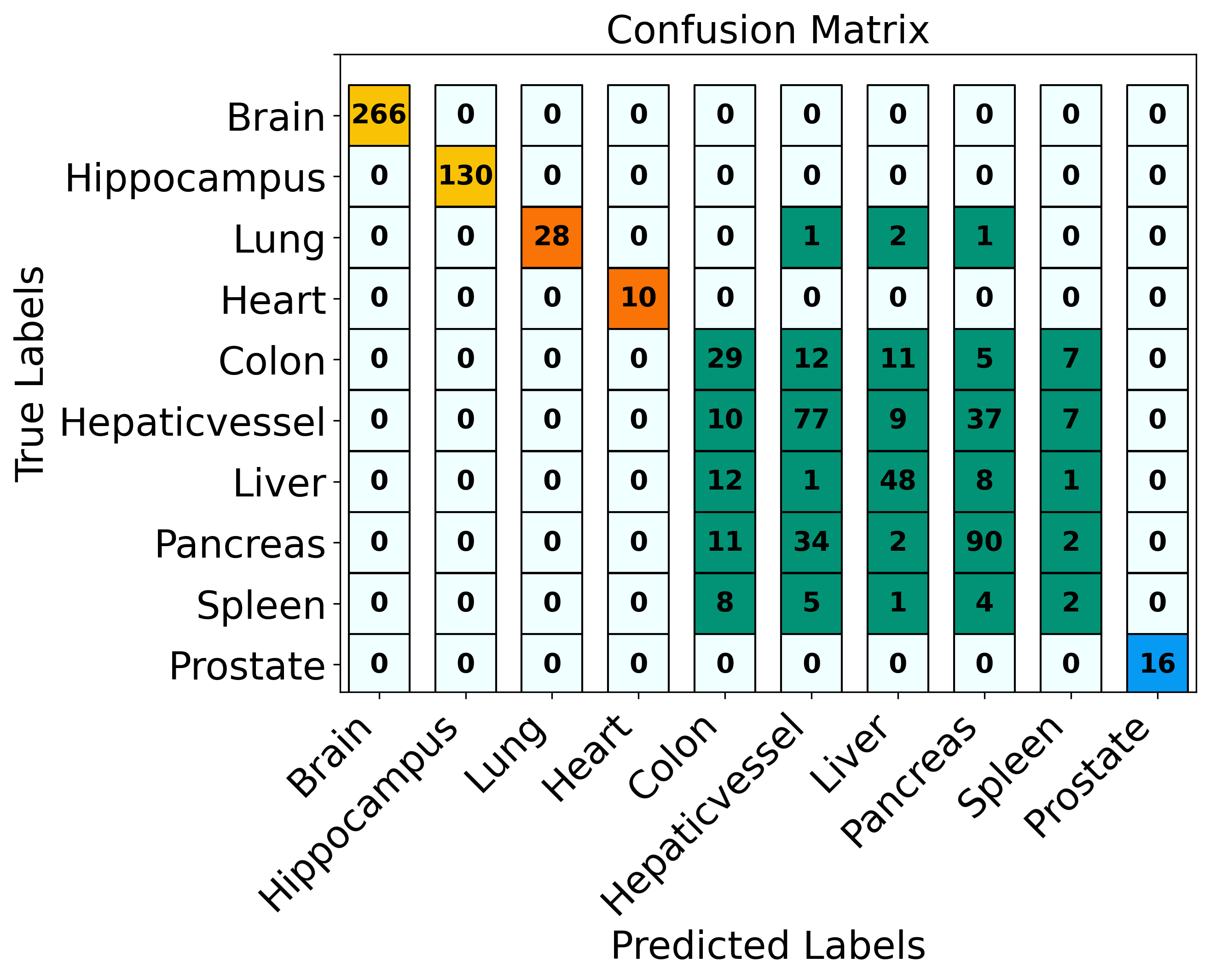}
        \caption{Organ level confusion matrix of the best method based on body region evaluation, colors represent the body regions: \textcolor[HTML]{FAC205}{Head}, \textcolor[HTML]{069AF3}{Pelvis}, \textcolor[HTML]{F97306}{Chest}, and \textcolor[HTML]{029376}{Abdomen}. \\ }
        \label{fig: CM figure}
    \end{minipage}
    \hfill 
    \begin{minipage}{0.48\textwidth}
        \centering
        \includegraphics[scale=0.22]{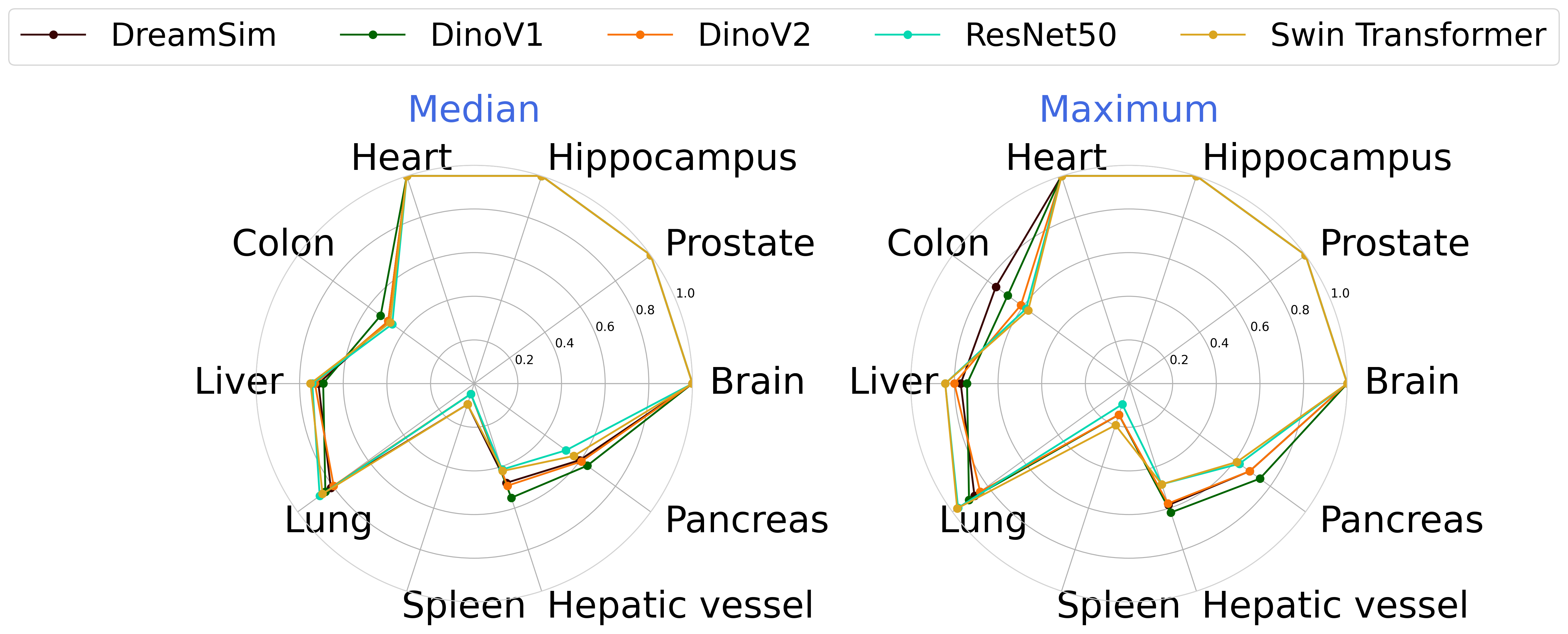}
        \caption{Organ level median (left) and maximum (right) of recall values across all variants of search indices and the aggregation schemes.}
        \label{fig:combined max-median}
    \end{minipage}
\end{figure}

\section{Conclusion}

We showed that without any retraining or fine-tuning pre-trained networks can be used for medical retrieval tasks even if they are trained on non-medical data for non-CBIR tasks. We could achieve perfect recall for modality and multiple body regions.   
The proposed method uses embeddings from pretrained models, computes per-slice embeddings only once, and combines the information from different slices using various sampling and weighting schemes which leverages the spatial redundancy in medical image volumes. 
The simple weighting might not be suitable for some organs or body regions and based on the information available on different target regions the weighting scheme should be adjusted. 
However, the spatial redundancy of the medical images can be leveraged by sampling the slices and therefore decreasing the size of the search space. 
Simple heuristic rules, e.g., slice rejection based on intensity values or height for specific organs can also be employed to improve the method which can be considered in future studies.  
LSH and HNSW indexing approaches perform on par on the tasks for different embeddings but since HNSW is significantly faster we recommend using this method. 

Using models trained on domain-specific medical images did not improve the retrieval. But we hesitate to conclude any superiority. 
There are certain limitations to point out. 
Firstly, the DINO-based models are in general larger networks with more parameters. Secondly, the Swin Transformer and ResNet50 are trained \emph{supervised} while the DINO-based models are all trained \emph{self-supervised}. 
Nevertheless, they perform on par. 
It is noteworthy that we used MSD labels where for each volume only single organs are labeled, thus, the organ-level performance does not include scenarios where multiple organs are visible in the images which occurs frequently, especially in the Abdomen region. Further studies should take this into account and more precise slice labels should be derived from segmentation masks. A possible solution can be using multiple labels per slice to compensate for the mentioned issue.

\section*{Compliance with Ethical Standards}

This research study was conducted retrospectively using human subject data made available in open access by \url{https://medicaldecathlon.com/}. Ethical approval was not required as confirmed by the license attached with the open-access data.

\section*{Acknowledgement}
We thank Timothy Deyer and his RadImageNet team for providing the RadImagenet pre-trained model weights for the Swin Transformer architecture.

\section*{Conflicts of Interest}
No funding was received for conducting this study. The authors have no relevant financial or non-financial interests to disclose.
    
\bibliographystyle{IEEEtran}
\bibliography{conference_101719}

\newpage

\section*{Appendix A}
\label{appendix-precision}

The precision values on the body region level are shown in \cref{tab:Body region precision} for all different combinations of search index, embedding, and aggregation scheme. 

\subsection{Modality Level}
Similar to the recall, on the modality level, for all possible combinations of the search index, embedding, and aggregation scheme a precision of $1.0$ was obtained. 

\subsection{Body Region Level}

\subsubsection{All Slices} The Head and Pelvis regions of all the models and indexing approaches have a perfect precision of $1.0$, except for DINOv2 and LSH indexing which has a slight drop to $0.99$. 
For the retrieval of the Abdomen region ResNet50 and LSH indexing has the highest precision of $0.997$. Swin Transformer is the second best with a precision of $0.993$. The DINO-based models have a slightly lower precision in comparison ($[0.880-0.990]$). 
For Chest, however, regardless of the indexing, DINOv1 performs the best with the precision of $1.0$. 

\subsubsection{Weighted Slices} A perfect precision of $1.0$ for the Head and Pelvis regions for all models is achieved. 
For the Abdomen region, the introduction of weighting decreased the precision of all models except for DINOv1 which achieved a precision of $1.0$. 
On the other hand, for the Chest region, the precision of all models except for ResNet50 increased. 

The increase and decrease of the precision values contrary to the drop of recall values are in a small range and for all the body regions all the models still have a precision higher than $0.9$.  

\subsubsection{Sampled Slices} 
Similar to the recall trends, subsampled slices in most cases perform on par with using all the slices in terms of their precision.

\begin{table*}[!htbp]
    \centering
     {\tiny
    \caption{Precision for different body regions Abdomen, Chest, Head, and Pelvis. Comparison: different vector database indexes (Id.), sampling strategies, aggregation variants all, per Image-id level and slice wise (SW)}
    \label{tab:Body region precision}
    \begin{tabular}{|p{0.2cm}|p{0.2cm}|p{0.8cm}|p{0.4cm}p{0.4cm}p{0.4cm}p{0.4cm}p{0.6cm}|p{0.4cm}p{0.4cm}p{0.4cm}p{0.4cm}p{0.4cm}p{0.6cm}|p{0.4cm}p{0.4cm}p{0.6cm}|p{0.6cm}|p{0.6cm}|p{0.6cm}|p{0.6cm}|} \hline
    \multirow{2}{*}{} & \multirow{2}{*}{Id.}  & \multirow{2}{*}{Embedding}  & \multicolumn{5}{|c|}{\# Random Samples}  & \multicolumn{6}{|c|}{\# Sampled every x mm} & \multicolumn{3}{|c|}{\# every n slices} & \multirow{2}{*}{all} & \multirow{2}{*}{W} & \multirow{2}{*}{SW}\\ \cline{4-17} 
         & & & 5 & 10 & 20 & 40 & 80 & 3 & 5 & 7 & 10 & 12 & 15 & 3 & 5 & 10 & & &\\ \hline \cline{1-17}
    \multirow{10}{*}{\RotTxt{Abdomen}}     & \multirow{5}{*}{\RotTxt{HNSW}} 
            &  DreamSim & .986 & .986 & .981 & .986 & .986 & .990 & .986  & .983 & .986 & .986 & .979 & .990 & .988 & .984 & .990 &  .960 & .956\\ 
         &  &  DINOv1 & .986 & .988 & .988 & .988 & .988 & .990 & .990 & .993 & .993 & .988 & .986 & .990 &.990  & .988 &  .990& .970 & .967\\
         &  &  DINOv2 & .981 & .983 & .988 & .986 & .988 & .986 & .988 & .986 & .988 & .983 & .983 & .986 & .988 & .988 &  .988& .964 &..956\\ 
         &  &  ResNet50 & .988 & .988 & .988 & .993 & .993 & .990 & .993 & .990 & .993& .990 & .990 & .990 & .995 & .993 & .990 & .972  & .956\\ 
         &  & SwinTrans.& .988 & .988 & .995 & .988 & .933 & .993 & .995 & .990 & .995 & .993 & .995  & .993 & .990 & .990 & .993 & .977 & .969\\ \cline{2-3}
         & \multirow{5}{*}{\RotTxt{LSH}} 
         &  DreamSim & .981 & .988 & .983 & .981 & .984 & .986 & .990 & .983 & .988 & .988 & .988 & .990 & .986 & .986 & .986 & .957 &.954\\
         &  &  DINOv1 & .981 & .986 & .988 & .990 & .988 & .988 & .988 & .990 & .986 & .988 & .986 & .988 & .990 & .986 & .990 & \textbf{1.0} &.961 \\
         &  &  DINOv2 & .974 & .984 & .983 & .983 & .983 & .923 & .923 & .925 & .921 & .945 & .945 & .984 & .988 & .984 & .986 & .953  & .952\\ 
         &  &  ResNet50 & .986 & .986 & .986 & .986 & .990 & .990 & .997 & .993 & .993 & .990 & .993 & .993 & .997 & .993 & .997 & .966 &  .956\\ 
         &  & SwinTrans. & .988 & .983 & .993 & .990 & .986 & .988 & .986 & .988 & .984 & .990 & .990 & .988 & .990 & .986 & .988 & .977 & .966\\ \hline
    \multirow{10}{*}{\RotTxt{Chest}}   & \multirow{5}{*}{\RotTxt{HNSW}}  
            &  DreamSim & .923 & .9 & .944 & .878 & .972 & .926 & .923 & .921 & .923 & .923 & .942 & .990 & .948 & .945 & .928 &  \textbf{1.0} & .882\\ 
         &  &  DINOv1 & .947 & \textbf{1.0} & \textbf{1.0} & \textbf{1.0} & .948 & .990 & .993 & .990 & .993 & .988 & .986 & .994& \textbf{1.0} & .973 & \textbf{1.0} &  \textbf{1.0}  & .920\\
         &  &  DINOv2 & .871 & .897 & .973 & .923 & .948 & .923 & .925 & .923 & .925 & .921 & .897 & .923 & .948 & .948 & .925 & \textbf{1.0}   & .859\\ 
         &  &  ResNet50 & .840 & .902 & .925 & .906 & .928 & .926 & .928 & .926 & .928 & .926 & .926 & .926 & .930 & .951 & .926 & .967  & .859 \\
         &  & SwinTrans. & .940 & .948 & .930 & .948 & .928 & .951 & .930 & .95 & .952 & .928 & .975 & .993 & .990 & .990 & .975 & .969  &.884 \\ \cline{2-3}
         & \multirow{4}{*}{\RotTxt{LSH}} &  DreamSim & .918 & .948 & .897 & .918 & .945 & .947 & .95 & .921 & .973 & .973 & .902 & .925 & .947 & .947 & .923 &  .958 & .873\\ 
         &  &  DINOv1 & .944 & \textbf{1.0} & .973 & .95 & \textbf{1.0} & \textbf{1.0} &  \textbf{1.0} & \textbf{1.0} & \textbf{1.0} & \textbf{1.0} & \textbf{1.0} & \textbf{1.0} & \textbf{1.0} & \textbf{1.0} & \textbf{1.0}  &  \textbf{1.0} & .909 \\
         &  &  DINOv2 & .861 & .945 & .921 & .875 & .921 & .923 & .923 & .925 & .921 & .945 & .945 & .945 & .948 & .945 & .923 & \textbf{1.0}  & .828 \\
         &  &  ResNet50 & .851 & .837 & .857 & .9 & .883 & .926 & .931 & .931 & .975 & .95 & .975 & .951  & \textbf{1.0} & .886 & .966  & .931  & .828 \\
         &  & SwinTrans.& .973 & .853 & .951 & .95 & .972 & .973 & .947 & .948 & .945 & .95 & .926 & .948 & .974 & .972 & .973 &  .969 & .890\\ \hline
    \multirow{10}{*}{\RotTxt{Head}}     & \multirow{5}{*}{\RotTxt{HNSW}} 
    &  DreamSim & \textbf{1.0} & \textbf{1.0} & \textbf{1.0}&\textbf{1.0} & \textbf{1.0}& .992 & .992 & .992 & .985 & .994 & .992 & \textbf{1.0}& \textbf{1.0}&\textbf{1.0}&\textbf{1.0}&\textbf{1.0}& .999 \\ 
         &  &  DINOv1 & .992 & .99 & .992 & .985 & .987 &\textbf{1.0} & .994 &\textbf{1.0} &\textbf{1.0} &\textbf{1.0} & \textbf{1.0}& \textbf{1.0}& \textbf{1.0}&\textbf{1.0}&\textbf{1.0}&\textbf{1.0}& .999 \\ 
         &  &  DINOv2 & \textbf{1.0} &\textbf{1.0}  & \textbf{1.0}& \textbf{1.0}&\textbf{1.0} & \textbf{1.0}& \textbf{1.0}& \textbf{1.0}& .997& \textbf{1.0}& \textbf{1.0}& \textbf{1.0}& \textbf{1.0}&.997&\textbf{1.0}&\textbf{1.0}& .999 \\  
         &  &  ResNet50 & .997 & .997 & .997 & .997 & .997 & .997 & .997 & .997 & .997 & .997 & .994 & .993 & .997 & .997 &\textbf{1.0} &\textbf{1.0} &  .999 \\ 
           &  & SwinTrans.& \textbf{1.0}& \textbf{1.0}& \textbf{1.0}& \textbf{1.0}&\textbf{1.0} & \textbf{1.0}&\textbf{1.0} &\textbf{1.0} &\textbf{1.0} &\textbf{1.0} &\textbf{1.0} &\textbf{1.0} &\textbf{1.0} &\textbf{1.0} &\textbf{1.0} &\textbf{1.0} &  .999  \\ \cline{2-3}
         & \multirow{5}{*}{\RotTxt{LSH}} &  DreamSim & \textbf{1.0} & \textbf{1.0} & \textbf{1.0}& \textbf{1.0}& \textbf{1.0}&\textbf{1.0} & \textbf{1.0}& \textbf{1.0}&\textbf{1.0} &\textbf{1.0} &\textbf{1.0} & \textbf{1.0}& \textbf{1.0}&\textbf{1.0}&\textbf{1.0}&\textbf{1.0}& .999 \\ 
         &  &  DINOv1 & \textbf{1.0} &\textbf{1.0}  &\textbf{1.0} &\textbf{1.0} &\textbf{1.0} & \textbf{1.0}& \textbf{1.0}&\textbf{1.0} & \textbf{1.0}&\textbf{1.0} & \textbf{1.0}&\textbf{1.0} & \textbf{1.0}&\textbf{1.0}&\textbf{1.0}&\textbf{1.0}& .999 \\
         &  &  DINOv2 & \textbf{1.0} &\textbf{1.0}  & \textbf{1.0}&\textbf{1.0} & \textbf{1.0}& \textbf{1.0}& \textbf{1.0}& \textbf{1.0}&\textbf{1.0} &\textbf{1.0} &\textbf{1.0} &\textbf{1.0} &\textbf{1.0} &\textbf{1.0}&.99 &\textbf{1.0}& .999 \\ 
         &  &  ResNet50 & .997 & .997 & .997 & .997 & .883 & .997 & .997 &\textbf{1.0} &\textbf{1.0} & \textbf{1.0} & .997 & .997 & .997 & .997 &\textbf{1.0} & \textbf{1.0} & .997 \\
         &  & SwinTrans.&\textbf{1.0} & \textbf{1.0}&\textbf{1.0} &\textbf{1.0} & \textbf{1.0}& \textbf{1.0}&\textbf{1.0} &\textbf{1.0} &\textbf{1.0} &\textbf{1.0} &\textbf{1.0} &\textbf{1.0} &\textbf{1.0} & \textbf{1.0}&\textbf{1.0} & \textbf{1.0} & .999 \\ \hline
    \multirow{10}{*}{\RotTxt{Pelvis}} & \multirow{5}{*}{\RotTxt{HNSW}} &  DreamSim &\textbf{1.0}  & \textbf{1.0} &\textbf{1.0} &\textbf{1.0} &\textbf{1.0} &\textbf{1.0} & \textbf{1.0}&\textbf{1.0} &\textbf{1.0} &\textbf{1.0} & \textbf{1.0}& \textbf{1.0}& \textbf{1.0}&\textbf{1.0}&\textbf{1.0}& \textbf{1.0}& \textbf{1.0}\\ 
         &  &  DINOv1 & .866 & .75 &.812 &.769 &.687 & .812& .812&.812 &.714 &.875 &.928 &.875 & \textbf{1.0}&\textbf{1.0}& \textbf{1.0} &\textbf{1.0}& .992\\ 
         &  &  DINOv2 & \textbf{1.0} & \textbf{1.0} &\textbf{1.0} &\textbf{1.0} & \textbf{1.0}& \textbf{1.0}& \textbf{1.0}& \textbf{1.0}& .941 & \textbf{1.0}&\textbf{1.0}  &\textbf{1.0} &\textbf{1.0} & .994 &\textbf{1.0}&\textbf{1.0}& \textbf{1.0}\\  
         &  &  ResNet50 & \textbf{1.0}& \textbf{1.0}& \textbf{1.0}& \textbf{1.0}& \textbf{1.0}& \textbf{1.0}&\textbf{1.0} &\textbf{1.0} &\textbf{1.0} &\textbf{1.0} &\textbf{1.0} & \textbf{1.0}& \textbf{1.0}&\textbf{1.0} &\textbf{1.0} & \textbf{1.0}& .996 \\
         &  & SwinTrans. & \textbf{1.0}& \textbf{1.0}& \textbf{1.0}& \textbf{1.0}& \textbf{1.0}& \textbf{1.0}&\textbf{1.0} &\textbf{1.0} &\textbf{1.0} &\textbf{1.0} &\textbf{1.0} & \textbf{1.0}& \textbf{1.0}&\textbf{1.0} &\textbf{1.0} & \textbf{1.0}& \textbf{1.0} \\ \cline{2-3}
         & \multirow{5}{*}{\RotTxt{LSH}}  &  DreamSim &  \textbf{1.0}& \textbf{1.0} & \textbf{1.0}&\textbf{1.0} &\textbf{1.0} & \textbf{1.0}&\textbf{1.0} &\textbf{1.0} &\textbf{1.0} & \textbf{1.0}&\textbf{1.0} &\textbf{1.0} &\textbf{1.0} &\textbf{1.0}&\textbf{1.0}&\textbf{1.0}& \textbf{1.0} \\ 
         &  &  DINOv1 & \textbf{1.0} & \textbf{1.0} &\textbf{1.0} &\textbf{1.0} &\textbf{1.0} & \textbf{1.0}& \textbf{1.0}&\textbf{1.0} &\textbf{1.0} &\textbf{1.0} &\textbf{1.0} &\textbf{1.0} & \textbf{1.0}&\textbf{1.0}&\textbf{1.0}&\textbf{1.0}& \textbf{1.0}\\ 
 & & DINOv2 &\textbf{1.0} & \textbf{1.0} &\textbf{1.0} &\textbf{1.0} &\textbf{1.0} &\textbf{1.0} & \textbf{1.0}& \textbf{1.0}& \textbf{1.0}& \textbf{1.0}& \textbf{1.0}& \textbf{1.0}&\textbf{1.0} &\textbf{1.0}&\textbf{1.0}&\textbf{1.0}&\textbf{1.0} \\
 &  &  ResNet50 &\textbf{1.0} &\textbf{1.0} & \textbf{1.0}&\textbf{1.0} & \textbf{1.0} &\textbf{1.0}&\textbf{1.0} &\textbf{1.0} &\textbf{1.0} &\textbf{1.0} &\textbf{1.0}& \textbf{1.0}& \textbf{1.0}& \textbf{1.0}& \textbf{1.0}&\textbf{1.0} & .952  \\ 
  &  &  SwinTrans. &\textbf{1.0} &\textbf{1.0} & \textbf{1.0}&\textbf{1.0} & \textbf{1.0} &\textbf{1.0}&\textbf{1.0} &\textbf{1.0} &\textbf{1.0} &\textbf{1.0} &\textbf{1.0}& \textbf{1.0}& \textbf{1.0}& \textbf{1.0}& \textbf{1.0}&\textbf{1.0} & \textbf{1.0}\\ \hline
\multicolumn{3}{|c|}{Overall average} & .968 &	.970 &	.976 & .971 & .970 & .977 & .976 &	.976 &	.974 &	.980 & .980 & .983 & .989 &	.985 &	.985 &	.987 &	.956 
 \\ \hline
    \end{tabular}
    }
\end{table*}

\section*{Appendix B}
\label{app:organ level}

\cref{tab: organ all} shows the detailed recall values at the organ level. The summarized statistics were shown in \cref{fig:combined max-median}.  

\begin{table*}[!t]
    \centering
    
    \caption{Recall for different organs. Comparison: different vector database indexes (Id.), sampling strategies, aggregation variants all, per Image-id level and slice wise (SW)}
    \label{tab: organ all}
    \tiny{
    \begin{tabular}{|p{0.2cm}|p{0.2cm}|p{0.8cm}|p{0.4cm}p{0.4cm}p{0.4cm}p{0.4cm}p{0.6cm}|p{0.4cm}p{0.4cm}p{0.4cm}p{0.4cm}p{0.4cm}p{0.6cm}|p{0.4cm}p{0.4cm}p{0.6cm}|p{0.6cm}|p{0.6cm}|p{0.6cm}|p{0.6cm}|} \hline
    \multirow{2}{*}{} & \multirow{2}{*}{Id.}  & \multirow{2}{*}{Embedding}  & \multicolumn{5}{|c|}{\# Random Samples}  & \multicolumn{6}{|c|}{\# Sampled every x mm} & \multicolumn{3}{|c|}{\# every n slices} & \multirow{2}{*}{all} & \multirow{2}{*}{W} & \multirow{2}{*}{SW}\\ \cline{4-17}
         & & & 5 & 10 & 20 & 40 & 80 & 3 & 5 & 7 & 10 & 12 & 15 & 3 & 5 & 10 & & &\\ \hline \cline{1-17}
    \multirow{10}{*}{\RotTxt{Brain}}& \multirow{5}{*}{\RotTxt{HNSW}} 
            &  DreamSim &  \textbf{1.0}& \textbf{1.0}& \textbf{1.0}& \textbf{1.0}& \textbf{1.0}& \textbf{1.0}& \textbf{1.0}& \textbf{1.0}& \textbf{1.0}& \textbf{1.0}& \textbf{1.0}&\textbf{1.0} & \textbf{1.0} &\textbf{1.0}& \textbf{1.0}& \textbf{1.0}&\textbf{1.0}\\ 
         &  &  DINOv1 &  \textbf{1.0}&  \textbf{1.0}&\textbf{1.0} & \textbf{1.0}& \textbf{1.0}& \textbf{1.0}&\textbf{1.0} & \textbf{1.0}& \textbf{1.0}& \textbf{1.0}& \textbf{1.0}& \textbf{1.0}& \textbf{1.0} &\textbf{1.0}& \textbf{1.0}&\textbf{1.0} &\textbf{1.0}\\
         &  &  DINOv2 & \textbf{1.0} &  \textbf{1.0}& \textbf{1.0}& \textbf{1.0}& \textbf{1.0}& \textbf{1.0}& \textbf{1.0}& \textbf{1.0}& \textbf{1.0}& \textbf{1.0}& \textbf{1.0}& \textbf{1.0}&  \textbf{1.0}& \textbf{1.0}& \textbf{1.0}&\textbf{1.0} &\textbf{1.0}\\  
         &  &  ResNet50 &\textbf{1.0} &\textbf{1.0} & \textbf{1.0} &\textbf{1.0} &\textbf{1.0} & \textbf{1.0} &\textbf{1.0} &\textbf{1.0} & \textbf{1.0} &\textbf{1.0} &\textbf{1.0} &\textbf{1.0} &\textbf{1.0} &\textbf{1.0} &  \textbf{1.0}&\textbf{1.0} &.999\\ 
         &  &  SwinTrans. &\textbf{1.0} &\textbf{1.0} & \textbf{1.0} &\textbf{1.0} &\textbf{1.0} & \textbf{1.0} &\textbf{1.0} &\textbf{1.0} & \textbf{1.0} &\textbf{1.0} &\textbf{1.0} &\textbf{1.0} &\textbf{1.0} &\textbf{1.0} &  \textbf{1.0}&\textbf{1.0} &.999 \\ \cline{2-3}
         & \multirow{5}{*}{\RotTxt{LSH}} 
            &  DreamSim & \textbf{1.0} & \textbf{1.0} & \textbf{1.0}& \textbf{1.0}& \textbf{1.0}&\textbf{1.0} &\textbf{1.0} &\textbf{1.0} &\textbf{1.0} &\textbf{1.0} & \textbf{1.0}& \textbf{1.0}&\textbf{1.0} & \textbf{1.0}& \textbf{1.0}& \textbf{1.0}&\textbf{1.0}\\
         &  &  DINOv1 & \textbf{1.0} & \textbf{1.0} & \textbf{1.0}& \textbf{1.0}& \textbf{1.0}& \textbf{1.0}& \textbf{1.0}& \textbf{1.0}& \textbf{1.0}& \textbf{1.0}& \textbf{1.0}& \textbf{1.0}&\textbf{1.0} & \textbf{1.0}& \textbf{1.0}&\textbf{1.0} &0.99\\
         &  &  DINOv2 & \textbf{1.0} & \textbf{1.0} &\textbf{1.0} &\textbf{1.0} &\textbf{1.0} & \textbf{1.0}& \textbf{1.0}& \textbf{1.0}& \textbf{1.0}&\textbf{1.0} &\textbf{1.0} & \textbf{1.0}&\textbf{1.0} & \textbf{1.0}& \textbf{1.0}&\textbf{1.0} &0.99\\ 
         &  &  ResNet50  & \textbf{1.0} &\textbf{1.0} &\textbf{1.0}  & \textbf{1.0} &\textbf{1.0} & \textbf{1.0} &\textbf{1.0} &\textbf{1.0}  & \textbf{1.0} &\textbf{1.0} &\textbf{1.0}  & \textbf{1.0}&\textbf{1.0} & \textbf{1.0}&  \textbf{1.0}& \textbf{1.0}&.999\\
         &  &  SwinTrans. &\textbf{1.0} &\textbf{1.0} & \textbf{1.0} &\textbf{1.0} &\textbf{1.0} & \textbf{1.0} &\textbf{1.0} &\textbf{1.0} & \textbf{1.0} &\textbf{1.0} &\textbf{1.0} &\textbf{1.0} &\textbf{1.0} &\textbf{1.0} &  \textbf{1.0}&\textbf{1.0} & \textbf{1.0} \\ \hline
    \multirow{10}{*}{\RotTxt{Colon}}   & \multirow{5}{*}{\RotTxt{HNSW}}  
            &  DreamSim &  .51&  .42& .45& .49& .5& .5&.515 & .515& .5& .515&.453 &.453 & .515 &.484 & .484&\textbf{.704} &.364\\ 
         &  &  DINOv1 &  .5& .491 & .531 &.484 &.5 & .546& .546& .562& .546& .531&.562 & .625 & .5&	.562 & .564&\textbf{.671} &.424\\
         &  &  DINOv2 &  .258& .421 &.5 &.5 & .5&.542 & .535& .507& .515& .531& .564& .531 & .484 & .564 & .468&\textbf{.590} &.324\\ 
         &  &  ResNet50 & .370& .453& .419& .393&.475 & .492& .492& .491& .492& .532& .476& .507&.523 &.387 & .484 &\textbf{.583} & .294\\ 
         &  &  SwinTrans. & .327 & .354 & .373 & .363 & .430 &  .539 & .539 & .539 &  .451 & .476 & .507 & .5 & .571 & .460 & .53 & .3 &  .332 \\ \cline{2-3}
         & \multirow{5}{*}{\RotTxt{LSH}}
            &  DreamSim & .442 &  .421& .507&.421 &.453 &.437 & .437& .437& .5& .515& .468&  .484 & .516 &	.546 & .421& \textbf{.754} &.358\\ 
         &  &  DINOv1 &  .606&  .546& .468& .516& .5& .468& .453& .453& .453& .514&.5 & .562 & .451 & .548 & .468& \textbf{.687} &.396\\
         &  &  DINOv2 & .295 & .359 &.387 &.5 &.451 & .483& .483& .438& .467& .451& .403& .578 & .491 & .419 & .468& \textbf{.612} &.287\\ 
         &  &  ResNet50 & .265& .425& .453& .390&.437 & .437& .437& .435& .515& .515& .468& .484& \textbf{.571}&.435 &  .437&.460 & .304\\
         &  &  SwinTrans. & .326 & .32 & .466  & .482 & .442 &  .467 & .451 & .451 & .532 & .523 & .523 & .548 & .539 & .428 & .453 & .365 & .316 \\ \hline
    \multirow{10}{*}{\RotTxt{Hepatic vessel}}     & \multirow{5}{*}{\RotTxt{HNSW}} 
            &  DreamSim &  .421& .428 & .5& .5& .51& .471&.471 & .464& .471& \textbf{.528} & \textbf{.528} & .521& .457& .478& .5& .321&.422\\ 
         &  &  DINOv1 &  .5&  \textbf{.621}& .571& .557& .585& .592& .578& .578& .578& .571& .557& .592 & .507 & .578 & .578& .471&.454\\ 
         &  &  DINOv2 &  .492&  .507& .557& \textbf{.578} & .492 & .542& .535& .507& .514& .5& .442& .478 & .435 & .492 & .542& .314&.421\\ 
         &  &  ResNet50 & .371& .414& .421& .435& .492& .471& \textbf{.478}& .457& .421& .385& .407&.414 &.4 &.364 & .471 &.292 &.374\\ 
         &  &  SwinTrans. & .524 & .57 &  .556 & .566 & .566 & .407  & .407 & .392 & .428  & .407 & .435 & .421 &  .421 & .464 & .407 & .589 & .371 \\ \cline{2-3}
         & \multirow{5}{*}{\RotTxt{LSH}}
            &  DreamSim &  .442 &  .421& .507& .478& .507& .471& .492& .442& .485& .478& .528& \textbf{.585} & .478 & .457 & .514& .307 &.423\\ 
         &  &  DINOv1 &  .514 &  .507&.55 &.55 &.557 & .535& .542& .55& .535& .514& .507& \textbf{.571} & .471 &	.542 & .542& .464 &.440\\
         &  &  DINOv2 &  .464 & .464 &.507 & \textbf{.564}& .542& .483& .483& .483& .467& .464&.507 & .464 & .457 & .55& .507& .335&.418\\ 
         &  &  ResNet50 & .357& .4& \textbf{.485}& .45&.421 & .442& .464& .414& .385& .357& .35& .407& .364&.414 & .428 &.242 &.341\\
         &  &  SwinTrans. & .479 & .512 & .578  & .539  & .565 &  .414 & .428 & .4 & .485  & .485 & .414 & .428 & .385 & .442 & .435 & .606 & .352 \\ \hline
    \multirow{10}{*}{\RotTxt{Hippocampus}} & \multirow{5}{*}{\RotTxt{HNSW}} 
            &  DreamSim &  \textbf{1.0}&  \textbf{1.0}& \textbf{1.0}& \textbf{1.0}& \textbf{1.0}& \textbf{1.0}& \textbf{1.0}& \textbf{1.0}& \textbf{1.0}& \textbf{1.0}&\textbf{1.0} &\textbf{1.0} & \textbf{1.0}& \textbf{1.0}& \textbf{1.0}& \textbf{1.0}&\textbf{1.0}\\ 
         &  &  DINOv1 &  .992&  \textbf{1.0}& \textbf{1.0}& \textbf{1.0}& \textbf{1.0}& \textbf{1.0}& \textbf{1.0}& \textbf{1.0}& \textbf{1.0}& \textbf{1.0}& \textbf{1.0}&\textbf{1.0} &\textbf{1.0} & \textbf{1.0}& \textbf{1.0}& \textbf{1.0}&.998\\ 
         &  &  DINOv2 &  \textbf{1.0}&  \textbf{1.0}&\textbf{1.0} & \textbf{1.0}& \textbf{1.0}& \textbf{1.0}& \textbf{1.0}& \textbf{1.0}& .992& \textbf{1.0}& \textbf{1.0}& .992 & \textbf{1.0} & \textbf{1.0} & \textbf{1.0}& \textbf{1.0}&.998\\ 
         &  &  ResNet50 &\textbf{1.0} &\textbf{1.0} & \textbf{1.0}& \textbf{1.0}&\textbf{1.0} &\textbf{1.0} &\textbf{1.0} &\textbf{1.0} &\textbf{1.0} &\textbf{1.0} &\textbf{1.0}  &\textbf{1.0} &\textbf{1.0} &\textbf{1.0} &  \textbf{1.0}& \textbf{1.0}&\textbf{1.0}\\
         &  &  SwinTrans. &\textbf{1.0} &\textbf{1.0} & \textbf{1.0}& \textbf{1.0}&\textbf{1.0} &\textbf{1.0} &\textbf{1.0} &\textbf{1.0} &\textbf{1.0} &\textbf{1.0} &\textbf{1.0}  &\textbf{1.0} &\textbf{1.0} &\textbf{1.0} &  \textbf{1.0}& \textbf{1.0}&\textbf{1.0}\\ \cline{2-3}
         & \multirow{4}{*}{\RotTxt{LSH}}  
            &  DreamSim &  \textbf{1.0}&  \textbf{1.0}&\textbf{1.0} &\textbf{1.0} & \textbf{1.0}& \textbf{1.0}&\textbf{1.0} & \textbf{1.0}&\textbf{1.0} & \textbf{1.0}&\textbf{1.0} & \textbf{1.0}& \textbf{1.0} & \textbf{1.0}& \textbf{1.0}& \textbf{1.0}&\textbf{1.0}\\ 
         &  &  DINOv1 &  \textbf{1.0}&  \textbf{1.0}&\textbf{1.0} &\textbf{1.0} & \textbf{1.0}& \textbf{1.0}&\textbf{1.0} & \textbf{1.0}&\textbf{1.0} & \textbf{1.0}&\textbf{1.0}& \textbf{1.0}& \textbf{1.0} & \textbf{1.0} & \textbf{1.0}& \textbf{1.0}&\textbf{1.0}\\ 
         &  & DINOv2 & \textbf{1.0}&  \textbf{1.0}&\textbf{1.0} &\textbf{1.0} & \textbf{1.0}& \textbf{1.0}& \textbf{1.0}& \textbf{1.0}& \textbf{1.0}& \textbf{1.0}& \textbf{1.0}& \textbf{1.0} & \textbf{1.0}& \textbf{1.0}& \textbf{1.0}& \textbf{1.0}&\textbf{1.0}\\  
         &  &  ResNet50 & \textbf{1.0}& \textbf{1.0}& \textbf{1.0}& \textbf{1.0}& \textbf{1.0}& \textbf{1.0}& \textbf{1.0}& \textbf{1.0}  & \textbf{1.0}& \textbf{1.0}& \textbf{1.0}  & \textbf{1.0}& \textbf{1.0}& \textbf{1.0} &  \textbf{1.0}& \textbf{1.0}& \textbf{1.0} \\
         &  &  SwinTrans. & \textbf{1.0}& \textbf{1.0}& \textbf{1.0}& \textbf{1.0}& \textbf{1.0}& \textbf{1.0}& \textbf{1.0}& \textbf{1.0}  & \textbf{1.0}& \textbf{1.0}& \textbf{1.0}  & \textbf{1.0}& \textbf{1.0}& \textbf{1.0} &  \textbf{1.0}& \textbf{1.0}& \textbf{1.0} \\ \hline
    \multirow{10}{*}{\RotTxt{Heart}} & \multirow{5}{*}{\RotTxt{HNSW}} 
            &  DreamSim &  \textbf{1.0}& \textbf{1.0} &\textbf{1.0} & \textbf{1.0}& \textbf{1.0}& \textbf{1.0}& \textbf{1.0}& \textbf{1.0}& \textbf{1.0}& \textbf{1.0}& \textbf{1.0}& \textbf{1.0}&\textbf{1.0} & \textbf{1.0} & \textbf{1.0}& \textbf{1.0}&\textbf{1.0}\\ 
         &  &  DINOv1 &  \textbf{1.0}&  \textbf{1.0}& \textbf{1.0}& \textbf{1.0}& \textbf{1.0}& \textbf{1.0}& \textbf{1.0}& \textbf{1.0}& \textbf{1.0}& \textbf{1.0}& \textbf{1.0}&\textbf{1.0} &\textbf{1.0} & \textbf{1.0}& \textbf{1.0}& \textbf{1.0}&.993\\ 
         &  &  DINOv2 &  \textbf{1.0}& \textbf{1.0} &\textbf{1.0} &\textbf{1.0} &\textbf{1.0} & \textbf{1.0}& \textbf{1.0}& \textbf{1.0}& \textbf{1.0}& \textbf{1.0}& \textbf{1.0} &\textbf{1.0} &\textbf{1.0} &\textbf{1.0}& \textbf{1.0}& \textbf{1.0}&\textbf{1.0}\\ 
         &  &  ResNet50 &\textbf{1.0} & \textbf{1.0}&\textbf{1.0} &\textbf{1.0} &\textbf{1.0} & \textbf{1.0} &\textbf{1.0} &\textbf{1.0} & \textbf{1.0} &\textbf{1.0} &\textbf{1.0}  &\textbf{1.0} &\textbf{1.0} &\textbf{1.0} & \textbf{1.0} &.9 & .985\\
         &  &  SwinTrans. &\textbf{1.0} & \textbf{1.0}&\textbf{1.0} &\textbf{1.0} &\textbf{1.0} & \textbf{1.0} &\textbf{1.0} &\textbf{1.0} & \textbf{1.0} &\textbf{1.0} &\textbf{1.0}  &\textbf{1.0} &\textbf{1.0} &\textbf{1.0} & \textbf{1.0} & \textbf{1.0} & \textbf{1.0} \\ \cline{2-3}
         & \multirow{5}{*}{\RotTxt{LSH}}  
            & DreamSim & \textbf{1.0} & \textbf{1.0} &\textbf{1.0} &\textbf{1.0} &\textbf{1.0} &\textbf{1.0} & \textbf{1.0}&\textbf{1.0} &\textbf{1.0} &\textbf{1.0} &\textbf{1.0} &\textbf{1.0} &\textbf{1.0} &\textbf{1.0}& \textbf{1.0}& \textbf{1.0}&\textbf{1.0}\\ 
         &  & DINOv1 &  \textbf{1.0}&  \textbf{1.0}& \textbf{1.0}&\textbf{1.0} &\textbf{1.0} &\textbf{1.0} & \textbf{1.0}& \textbf{1.0}&\textbf{1.0} & \textbf{1.0}&\textbf{1.0} & \textbf{1.0} &\textbf{1.0} &\textbf{1.0} & \textbf{1.0}& \textbf{1.0}&\textbf{1.0}\\ 
         &  & DINOv2 & \textbf{1.0}&  \textbf{1.0}& \textbf{1.0}& \textbf{1.0}& \textbf{1.0}&\textbf{1.0} & \textbf{1.0}& \textbf{1.0}& \textbf{1.0}& \textbf{1.0}&\textbf{1.0} &\textbf{1.0} &\textbf{1.0} &\textbf{1.0}& \textbf{1.0}& \textbf{1.0}&\textbf{1.0}\\ 
         &  &  ResNet50 & \textbf{1.0}& \textbf{1.0}& \textbf{1.0}&\textbf{1.0} & \textbf{1.0}& \textbf{1.0}& \textbf{1.0}& \textbf{1.0}  & \textbf{1.0}& \textbf{1.0}& \textbf{1.0}  &  \textbf{1.0}& \textbf{1.0} &\textbf{1.0}& \textbf{1.0} &\textbf{1.0} &.978\\
         &  &  SwinTrans. & \textbf{1.0}& \textbf{1.0}& \textbf{1.0}&\textbf{1.0} & \textbf{1.0}& \textbf{1.0}& \textbf{1.0}& \textbf{1.0}  & \textbf{1.0}& \textbf{1.0}& \textbf{1.0}  &  \textbf{1.0}& \textbf{1.0} &\textbf{1.0}& \textbf{1.0} &\textbf{1.0} & \textbf{1.0} \\ \hline
    \multirow{10}{*}{\RotTxt{Liver}} & \multirow{5}{*}{\RotTxt{HNSW}} 
            &  DreamSim & .671 &  .671& .614& .685&.714 &.714 & .714& .7& .728&.7 & \textbf{.771} &.757 & .685 &.685& .728& .5&.658\\ 
         &  &  DINOv1 &  .614&  .628& .642& \textbf{.742}& .671& .7& .7& .685& \textbf{.742}& .714& .628 & .685 & .714 & .671 & .7& .471&.715\\ 
         &  &  DINOv2 & .614 & .7 & .685& .714& .728& .757& .742& .742& .728& .714& .757& .671 & .728 & \textbf{.771} & .742& .485&.674\\ 
         &  &  ResNet50 & .557& .585& .671& .7& .742& .757& .742& \textbf{.785}& .742& .714&.742 & .714& .714&.771 &  .757& .557&.608\\ 
         &  &  SwinTrans. & .455 & .560 &  .552 & .556 & .566 & .757 & .771 & .771 & .814  & .757 & .728 & .742 & .742 & .685 & .757 & .380 & .684 \\ \cline{2-3}
         & \multirow{5}{*}{\RotTxt{LSH}} 
            &  DreamSim & .628 &  .6& .685& .742&.685 & .742& \textbf{.757}& \textbf{.757}& .742& .728& .742& .7 & .714 & .742 & .742& .571&.661\\ 
         &  &  DINOv1 &  .542&  .714&.657 & .7&.714 & .714& .714& .685& .7&.685 &.671 &.7& .7 & .685 & \textbf{.728}& .514&.721\\ 
         &  &  DINOv2 & .614&  .642& .771& .714&.757 & .771& .757& .757& .728& .742& .671 & .757 & \textbf{.8} & .7& .771& .514&.677\\ 
         &  &  ResNet50 & .557& .614& .742& .742& .814& .8& .771& .814& .771& .771& .742&.742 & \textbf{.842}&.771 &  .8&.585 &.658\\
         &  &  SwinTrans. & .471 & .536 &  .515 & .563 & .509 &  .8 & .8 & .771 &  .828 & .842 & .785 & .8 & .8 & .7 & .8  & .381 & .693 \\ \hline
    \multirow{10}{*}{\RotTxt{Lung}} & \multirow{5}{*}{\RotTxt{HNSW}} 
            &  DreamSim &  .812&  .812& .75 & .812&.812 & \textbf{.875}& .812&.781 & .8& .812& .718& \textbf{.875}& .843 & .781& \textbf{.875}& .483&.700\\ 
         &  &  DINOv1 &  .812&  .843& .843& .843&.843 & .875& .875& \textbf{.906}& \textbf{.906}& .843& .812& .875 & .843 &	.875 & .875& .612&.776\\ 
         &  &  DINOv2 & .75 &  .781& \textbf{.843}& .812& \textbf{.843}&  .812&\textbf{.843} &.812 & \textbf{.843}& .781& .781 & \textbf{.843} & \textbf{.843} & .812 & \textbf{.843}& .615&.705\\ 
         &  &  ResNet50 & .843& .843& .843& .906&.906 & .875& .906& .875& .906&.875 &.875 &.875 &\textbf{.937} &.906 & .875 &.625 &.708\\
         &  &  SwinTrans. & .794 & .931 &  .909 & .931 & .906 & .906  & .937 & .875 & .937  & .906 & .937 & .906 & .875 & .875 & .906 & .956 &  .796\\ \cline{2-3}
         & \multirow{5}{*}{\RotTxt{LSH}} 
            & DreamSim &  .75& \textbf{.843} &.781 & .75& .781& .812& .875& .781& \textbf{.843}&\textbf{.843} & \textbf{.843}& \textbf{.843} & .812 &.812 & .812& .406&.685\\ 
         &  & DINOv1 &  .75 &  .838& .843& \textbf{.875}& .843& .843& .843& \textbf{.875}& .812& .843& .812&\textbf{.875} & .838 & .843& \textbf{.875}& .612 &.734\\ 
         &  & DINOv2 & .656 &  .781& .781&.781 & .781& .812& .812& \textbf{.843}& .781& .781& .781& .781 & \textbf{.843} & .781 & .812& .392& .679\\ 
         &  &  ResNet50 & .812& .812& .812& .812& .875&.875 & \textbf{.968}& \textbf{.968}& .906& .875& .906& .906& \textbf{.968}&.906 & \textbf{.968} &.566 &.707\\
         &  &  SwinTrans. & .964 & .806 & .935  & .933 & .962 &  .834 & .812 & .843 & .781 & .875 & .875 & .843 & .875 & .812 & .843 & .916 &  .775 \\ \hline
    \multirow{10}{*}{\RotTxt{Pancreas}} & \multirow{5}{*}{\RotTxt{HNSW}} 
            &  DreamSim &  .647&  .582& .611& .589&.604 & .604& .597& .597&.580 & .589& 0625& .618&.633 &.625& .604& \textbf{.683}&.582\\ 
         &  &  DINOv1 &  .618& .604 & .654& .633& .611& .654& .647& .647& .618& .625& .611& .611 & .640 & .633 & .654& \textbf{.741}&.602\\ 
         &  &  DINOv2 &  .604&  .611& .589& .611& .561& .582& .589& .597& .625& .618&.539 & .640 & .611 & .611& .582& \textbf{.683} & .557\\ 
         &  &  ResNet50 & .519&.489 & .517& .611&.604 & .575& .618& \textbf{.625}& .633& .618& .517 & .589&.561 & .496& .575 &.496 &.491\\ 
         &  &  SwinTrans. & .493 & .556 & .55  & .592 & .592 & .503 & .568 & .561 &  .539 & .568 & .589 & .532 & .589 & .604 & .510 & .492 &  .526 \\ \cline{2-3}
         & \multirow{5}{*}{\RotTxt{LSH}} 
            &  DreamSim & .517 & .604 & .568& .561& .561& .582& .561& .553& .539& .517& .633 & .633 & .611 & .589 & .546& \textbf{.669}&.553\\ 
         &  &  DINOv1 &  .618& .589 & .633 & .683&.654 & .640& .647& .647& .604&.615 &.625 & .654 & .640 & .654 & .640& \textbf{.726} &.604\\ 
         &  &  DINOv2 & .525&  .568& .611&.625 & .597 &.582 & .589&.597 & .618& .618& .625& .575 & \textbf{.654} & .640 & .582& \textbf{.654}&.555\\ 
         &  &  ResNet50 & .446& .424& .510& .525&.503 &.478 & .489& .489& .503& .503& .510&\textbf{.582} &.503 &.453 &  .474& .460&.455\\
         &  &  SwinTrans. & .476 & .578 & .590 & .551 & .569 &  .510 & .575 & .575 & .575  & .575 & .561 & .525 & .568 & .568 & .510 & .530 & .520  \\ \hline
    \multirow{10}{*}{\RotTxt{Prostate}} & \multirow{5}{*}{\RotTxt{HNSW}} 
            &  DreamSim & \textbf{1.0} &  \textbf{1.0}& \textbf{1.0}& \textbf{1.0}& \textbf{1.0}& \textbf{1.0}&\textbf{1.0} & \textbf{1.0}& \textbf{1.0}& \textbf{1.0}& \textbf{1.0}& \textbf{1.0} & \textbf{1.0} & \textbf{1.0} & \textbf{1.0}& .937&.977\\ 
         &  &  DINOv1&  .866& .75 & .812& .769&.687 & .812& .812& .812& .714& .875& .928& .875& \textbf{1.0}& \textbf{1.0}&  .733 &  \textbf{1.0}&.846\\ 
         &  &  DINOv2 &  \textbf{1.0}&\textbf{1.0}  &\textbf{1.0} &\textbf{1.0} & \textbf{1.0}& \textbf{1.0}& \textbf{1.0}&\textbf{1.0} & \textbf{1.0}&\textbf{1.0} &\textbf{1.0} &\textbf{1.0} & \textbf{1.0}&\textbf{1.0} & \textbf{1.0}& \textbf{1.0}&.980\\ 
         &  &  ResNet50 &\textbf{1.0} &\textbf{1.0} &\textbf{1.0} &\textbf{1.0} & \textbf{1.0}& \textbf{1.0}& \textbf{1.0}& \textbf{1.0}& \textbf{1.0}& \textbf{1.0}& \textbf{1.0} & \textbf{1.0}& \textbf{1.0}& \textbf{1.0}&\textbf{1.0}  &\textbf{1.0} &.979\\ 
         &  &  SwinTrans. &\textbf{1.0} &\textbf{1.0} &\textbf{1.0} &\textbf{1.0} & \textbf{1.0}& \textbf{1.0}& \textbf{1.0}& \textbf{1.0}& \textbf{1.0}& \textbf{1.0}& \textbf{1.0} & \textbf{1.0}& \textbf{1.0}& \textbf{1.0}&\textbf{1.0}  &\textbf{1.0} & \\ \cline{2-3}
         & \multirow{5}{*}{\RotTxt{LSH}}
            &  DreamSim & \textbf{1.0} &  \textbf{1.0}& \textbf{1.0}& \textbf{1.0}& \textbf{1.0}& \textbf{1.0}& \textbf{1.0}& \textbf{1.0}& \textbf{1.0}& \textbf{1.0}& \textbf{1.0}& \textbf{1.0} & \textbf{1.0} & \textbf{1.0} & \textbf{1.0}& \textbf{1.0}&.986\\ 
         &  &  DINOv1 &  \textbf{1.0}&  \textbf{1.0}& \textbf{1.0}&\textbf{1.0} &\textbf{1.0} & \textbf{1.0}&\textbf{1.0} &\textbf{1.0} & \textbf{1.0}& \textbf{1.0}& \textbf{1.0}& \textbf{1.0} & \textbf{1.0} & \textbf{1.0} & \textbf{1.0}& \textbf{1.0}&.986\\ 
         &  & DINOv2 & \textbf{1.0}& \textbf{1.0} & \textbf{1.0}&\textbf{1.0} &\textbf{1.0} & \textbf{1.0}& \textbf{1.0}&\textbf{1.0} & \textbf{1.0}& \textbf{1.0}& \textbf{1.0}& \textbf{1.0}& \textbf{1.0}&\textbf{1.0}& \textbf{1.0}& \textbf{1.0}&.986\\ 
         &  &  ResNet50 &\textbf{1.0} & \textbf{1.0}&\textbf{1.0} &\textbf{1.0} &\textbf{1.0} & \textbf{1.0}& \textbf{1.0}& \textbf{1.0} & \textbf{1.0}& \textbf{1.0}& \textbf{1.0} & \textbf{1.0}& \textbf{1.0}& \textbf{1.0} & \textbf{1.0} & \textbf{1.0}&.986\\ 
         &  &  SwinTrans. &\textbf{1.0} &\textbf{1.0} &\textbf{1.0} &\textbf{1.0} & \textbf{1.0}& \textbf{1.0}& \textbf{1.0}& \textbf{1.0}& \textbf{1.0}& \textbf{1.0}& \textbf{1.0} & \textbf{1.0}& \textbf{1.0}& \textbf{1.0}&\textbf{1.0}  &\textbf{1.0} & \\ \hline
    \multirow{10}{*}{\RotTxt{Spleen}} & \multirow{5}{*}{\RotTxt{HNSW}} 
            &  DreamSim &  .1&  0& \textbf{.15}& .05&.05 & .1& .1& .1& .05& .05& .1& .1 & .1 & .1 & .1& .1&.056\\ 
         &  &  DINOv1 & .1 & .1 & .05& .05&.05 & .05& .05& .05& \textbf{.15}& \textbf{.15}& .1& .05	& .05 & .1 & .05& .05&.047\\ 
         &  &  DINOv2 &  .05 & .05 & .1& .05& .05& .05& .05& .05 & .05& .05& .05 & \textbf{.15} & .1	& .05 & .05& .05&.057\\ 
         &  &  ResNet50 & .05& .05& 0& .05&.05 & .05& .05& .05&.05 & .05&.05 &.05 &0 &\textbf{.1} & .05 &.05 &.049\\ 
         &  &  SwinTrans. & 0 & .060 & .115 & .035 & .133 & .1 & .1 & .1 & .15 & .15 & .15 & .15 & .05 & 0 & .1 & 0 & .065 \\ \cline{2-3}
         & \multirow{5}{*}{\RotTxt{LSH}}
            &  DreamSim &  .05&  0& .05& 0& .1& .1& .1&.1 & .1& .1& .05& .05 & .05 & .1 & \textbf{.15}& 
            \textbf{.15}&.057\\ 
         &  &  DINOv1 &  .1&.05  &0 & .1& .1& .1& .1& .1& .05& .05&\textbf{.15} & 0 &.05 & \textbf{.15} & .05& .05&.049\\ 
         &  & DINOv2 & .05&  0& .05& .05&.05 & .05& .05& .05& 0& 0& \textbf{.1}& .05 & .05 & \textbf{.1} & 0& 0&.053\\ 
         &  &  ResNet50 & 0& \textbf{.15}& .05& .05&0 &0 &0 & 0& .05& .05& .1& .1& .05&.1 & 0 & \textbf{.15}&.049\\
         &  &  SwinTrans. & .083 & .071 &  .105 & .08 & .103 & .05 & .05 & .05 & .1 & .1 & .1 & .1 & .1 & .05 & .05 & .038 & .653 \\ \hline
        
    \end{tabular}
    }
    
\end{table*}

\end{document}